\documentclass[journal,twoside,web]{ieeecolor}

\usepackage{generic}
\usepackage{cite}
\usepackage{amsmath,amssymb,amsfonts}
\usepackage{algorithm}
\usepackage{algpseudocode}
\usepackage{graphicx}
\usepackage{textcomp}
\usepackage{threeparttable}
\usepackage{booktabs}
\usepackage{hyperref}
\usepackage{lcsys}

\newcommand{\name}{ASIA}

\begin{document}

\def\BibTeX{{\rm B\kern-.05em{\sc i\kern-.025em b}\kern-.08em
    T\kern-.1667em\lower.7ex\hbox{E}\kern-.125emX}}
\markboth{\journalname, VOL. XX, NO. XX, XXXX 2017}
{Piga and Forgione: \name{}: an AI Agent for System Identification}

\title{\name{}: an Autonomous System Identification Agent}

\author{
Dario Piga$^{a}$,
Marco Forgione$^{a}$%
\thanks{$^{a}$ Dalle Molle Institute for Artificial Intelligence (IDSIA), SUPSI, CH 6928 Lugano-Viganello, Switzerland (e-mail: \{dario.piga, marco.forgione\}@supsi.ch).}%
\thanks{This work was supported as a part of NCCR Automation, a National Centre of Competence in Research, funded by the Swiss National Science Foundation (grant number 51NF40\_225155).}
}

\maketitle
\thispagestyle{empty}

\begin{abstract}
Over the years, research in system identification has provided a rich set of
methods for learning dynamical models, together with well-established
theoretical guarantees.
In practice, however, the choice of model class, training algorithm, and
hyperparameter tuning is still largely left to empirical trial-and-error,
requiring substantial expert time and domain experience.
Motivated by recent advances in agentic artificial intelligence, we present
\emph{\name{}}, a framework that delegates this iterative search to a
large language model acting as an autonomous coding agent.
Building on existing agentic platforms, \name{} closes the loop between hypothesis, implementation,
and  evaluation without human intervention, requiring only
a plain-English description of the identification problem.
We conduct an empirical study of \name{} on two system identification
benchmarks and analyse the agent's search behaviour, the architectures and training
strategies it discovers, and the quality of the resulting models.
We also discuss the  potential of the approach and its current
limitations, including implicit test leakage, reduced methodological
transparency, and reproducibility concerns.
\end{abstract}


\section{Introduction}
\label{sec:introduction}
A system identification pipeline requires a set of interconnected design choices.
Given a collection of input-output trajectories, the practitioner must first
select a model structure, which may range from linear to nonlinear, time-invariant or time-varying, physical grey-box representations,
black-box neural networks, or hybrids thereof.
Architectural hyperparameters must then be specified: dynamical order, kernel functions  in kernel-based approaches,
and in neural networks the number of layers and neurons, the type of
recurrent cell (vanilla RNN, LSTM, GRU),  the use of residual connections
or regularisation, just to cite a few. 
Finally, optimisation hyperparameters must be tuned: cost function,
learning rate, regularisation strength, batch size, number of training epochs. 

In practice, these choices are still largely based on empirical decisions made by human experts, relying on domain expertise and prior experience. Partial automation is possible within a fixed model family. Random
search and Bayesian optimisation can explore continuous hyperparameter spaces without expert guidance, but the choice of model class and the design of the training algorithm itself typically remain human responsibilities. Although these tools have been successfully applied in identification and control~\cite{Sorourifar21,rath2024}, any change in the human-selected components generally requires re-tuning the associated hyperparameters. 
Overall, the process remains time-consuming and heavily reliant on domain expertise.

A complementary paradigm has emerged in the  AI  field, where recent works have explored the use of autonomous \emph{agents} based on Large Language Models (LLMs) to plan and execute empirical research and experimentation cycles~\cite{Lu2024,Huang24,Karpathy2026}.
The common paradigm is that of an LLM-based coding agent operating in an iterative loop with minimal or no human intervention. At each iteration, the agent reads a task description together with the history of previous experiments, proposes and implements modifications as executable code, runs the resulting program, and observes the new outcome.

Motivated by these new advances in AI,  the present work addresses the following question: \emph{Can an AI agent autonomously conduct the full system identification pipeline?}
To investigate this question, we present \emph{\name{}} (\emph{Autonomous System Identification Agent}), an agentic AI framework for system identification. 
Rather than introducing a new identification methodology, this paper provides an empirical study of how such an agent behaves
on common system identification benchmarks. Based on the \emph{autoresearch} platform recently proposed in~\cite{Karpathy2026},  \name{}  receives a plain-English
description of the identification problem and a  fixed evaluation protocol; it autonomously searches over model classes, architectures, and training strategies without further human intervention. Optionally, the agent can be provided with domain knowledge,  references to relevant literature, and candidate modelling strategies or architectures. 

A related line of work is meta-learning, which learns priors, initialisations, or update rules from a family of tasks to speed up adaptation on a new system~\cite{FoPuPi23,Zhe23}. In this sense, meta-learning and \name{} are complementary:
meta-learning aims to learn how to adapt, whereas \name{} searches for what to adapt.

We evaluate \name{} on two  benchmarks. 
The first one is the well-established \emph{Cascaded Two Tank} system~\cite{Schoukens2016}, that allows direct comparison with published results.
The second concerns identification of  \emph{nanodrone dynamics}~\cite{Busetto26}, a challenging multi-input multi-output problem where the agent can exploit physical domain knowledge.
In both cases, we analyse the agent's search behaviour and the solutions it discovers.  

We emphasise that our goal  is not to fully automate the field of system identification, but rather to explore how agentic AI can support and augment the model learning process. Based on our experience, we aim to critically assess  potential and   limitations of \name{} (and more in general, Agentic AI in research), highlighting strengths, failure modes,
and potential risks, while providing our transparent evaluation of its capabilities.

\section{\name{}: an Agentic Framework for System Identification}
\label{sec:method}

Let $\mathcal{D} = \{(u_k, y_k)\}_{k=1}^{T}$ denote a collection of input-output trajectories from an unknown dynamical system, partitioned into a training set $\mathcal{D}_{\mathrm{tr}}$ and a held-out test set
$\mathcal{D}_{\mathrm{te}}$.
A \emph{system identification configuration}~$\theta$ comprises, among others, the choice of model class $\mathcal{M}$ (e.g., state-space, neural networks,  physics-based ODE, hybrid), its architectural hyperparameters (e.g., number of states,  number of layers in a neural network, dropout
rate), and the training procedure (e.g., loss function, learning rate, 
regularisation coefficients, optimisation schedule).
Given a $K$-fold cross-validation protocol that partitions $\mathcal{D}_{\mathrm{tr}}$
into training and validation subsets $\{(\mathcal{D}^{(k)}_{\mathrm{tr}},\mathcal{D}^{(k)}_{\mathrm{val}})\}_{k=1}^{K}$,
the cross-validation metric is
\begin{equation}
  V(\theta)
  = \frac{1}{K}\sum_{k=1}^{K}
    \mathcal{L}\!\left(\hat{y}(\theta,\mathcal{D}^{(k)}_{\mathrm{tr}}),\,
                         \mathcal{D}^{(k)}_{\mathrm{val}}\right),
  \label{eq:cv_metric}
\end{equation}
where $\hat{y}(\theta,\mathcal{D}^{(k)}_{\mathrm{tr}})$ denotes the
multi-horizon predictions of a model trained with configuration~$\theta$
on fold~$k$, and $\mathcal{L}$ is the evaluation loss.
The system identification pipeline seeks
\begin{equation}
  \theta^{\star} = \mathop{\arg\min}_{\theta \in \Theta}\; V(\theta),
  \label{eq:sysid_opt}
\end{equation}
where the configuration space~$\Theta$ is combinatorial and includes model class,
architecture, and training procedure. These  are interdependent, and their joint
space can not be explored exhaustively within a realistic computational budget.

The new framework  \emph{\name{}} proposed in this paper uses a large language model which acts as the optimisation engine. At each
iteration it reads the current implementation and the full history of
previous trials~$\mathcal{T}$, proposes a new configuration~$\theta_k$,
implements and executes it, evaluates $V(\theta_k)$, and updates 
$\theta$ accordingly.

Following the \emph{autoresearch} framework~\cite{Karpathy2026}, the \emph{\name{}}  setup 
comprises the following components:
\begin{itemize}

\item A \textbf{problem description} document describing the identification
task, available data, prior physical knowledge, candidate
modelling approaches, and the protocol governing the search process,
including which components may be modified and which metrics drive model
selection.

\item A \textbf{read-only preprocessing and evaluation pipeline}, responsible for  generating the training and validation splits, and
computing the evaluation metric $V(\theta)$.
Keeping this pipeline immutable guarantees that all candidate solutions
are evaluated under an identical protocol and that reported metrics are
directly comparable across iterations.

\item A set of \textbf{modifiable modelling and training components} $\theta$,
including model architectures, loss functions, identification and
optimisation strategies, and related
hyperparameters, which can be freely edited by the agent.

\item An \textbf{iterative experimentation loop} in which the agent
analyses the results $\mathcal{T}$ of previous runs together with free-text
experiment descriptions recording the rationale behind past
modifications. Based on this information, the agent formulates a new
hypothesis, edits the relevant code, executes training and evaluation,
and records the resulting metrics, continuing until a computational
budget is exhausted or validation performance saturates.

\end{itemize}

In our experimental analysis, we instantiate the \emph{\name{}}  framework using Claude
Code 
powered by the Sonnet~4.6 LLM.

\section{Empirical Case Studies}
\label{sec:benchmarks}
In this section, we demonstrate the \name{}  framework on two system identification benchmarks: a cascaded two-tank
system~\cite{Schoukens2016} and  a brushless nanodrone~\cite{Busetto26}. 
In both benchmarks we follow the code structure of \emph{autoresearch}~\cite{Karpathy2026}. The agent is provided with a Markdown document describing, in plain English, the benchmark problem, available signals, candidate modelling approaches (including suggestions to explore physics-based and hybrid
architectures), and the search protocol (modifiable files, evaluation
metric, logging rules). Summaries extracted from the benchmark descriptions~\cite{Schoukens2016,Busetto26} are also provided to inject relevant domain knowledge.

$K$-fold cross-validation is adopted as the evaluation protocol. To this aim,  the
available training data are partitioned into trajectories used for
training and for validation, and the aggregated cross-validation error
is the sole criterion for model selection. A simple recurrent network
with plain gradient-based optimisation serves as the baseline
architecture for both case studies. The agent is free to modify the
model architecture, training strategy, optimisation algorithm, and
hyperparameters. A maximum training time per run is enforced to keep
the search computationally tractable. Furthermore, to promote diversity
in the early phase of the search, the protocol explicitly instructs the
agent to explore substantially different model families during the
first  iterations (spanning, for example, black-box, physics-based, and
hybrid architectures) before switching to refinements of the
most promising candidates.

The \name{} framework is available at~\url{https://github.com/dariopi/ASIA} and can be easily adapted to other identification problems. The framework can be run with any agentic coding environment, such as Claude Code or OpenAI Codex. For both benchmarks, the best model architecture selected by \name{} is available in the GitHub repository, and the corresponding training  procedure can be fully reproduced.

\subsection{Cascaded two-tank system}

\subsubsection{Benchmark description}
\noindent
The system to be identified consists of two water tanks connected in series.
A pump feeds water from a reservoir into the upper tank, which drains into the
lower tank through a small opening. The lower tank then drains back to the
reservoir through a second opening.
The system input $u$ is the water inflow in the upper tank, and the output $y$ is
the water level of the lower tank. The upper tank level is not measured.
Denoting the water levels of the upper and lower tank as $x_1$ and $x_2$
respectively, the basic system dynamics can be  approximated  by the nonlinear ODEs
\begin{equation}
    \left\{
    \begin{aligned}
        \dot{x}_1(t) &= -k_1\sqrt{x_1(t)} + k_4\, u(t) \\
        \dot{x}_2(t) &=  k_2\sqrt{x_1(t)} - k_3\sqrt{x_2(t)}
    \end{aligned}
    \right.
    \label{eq:tank_ode}
\end{equation}
where $k_1,k_2, k_3, k_4$ are unknown physical parameters related to the outlet
cross-sections and the pump gain.
In addition, a hard state saturation nonlinearity not captured by~\eqref{eq:tank_ode} arises when the tanks are completely full. This also induces an input-dependent process disturbance due to water overflowing from the upper to the lower tank.

Both the training and test sequences consist of $1024$ samples, collected at a
sampling period of $T_s = 4$\,s.
Following the benchmark guidelines~\cite{Schoukens2016}, performance is
evaluated by the RMSE of the open-loop simulation on the test set:
\begin{equation}
    \mathrm{RMSE} = \sqrt{\frac{1}{N}\sum_{k=k_0}^{K}\bigl(y_k-\hat{y}_k\bigr)^2},
    \label{eq:rmse}
\end{equation}
where $\hat{y}_k$ is the model prediction at time step $k$, $K = 1024$,
$k_0 = 51$, and $N = K - k_0 + 1 = 974$.
The first $50$ samples of the test sequence are discarded to eliminate transient
effects due to initial-condition estimation.

\subsubsection{Evaluation criterion}
\noindent
To guide the search process within the agentic framework, we adopted
a 2-fold cross-validation strategy on the available training trajectory.
The $1024$-sample training sequence is partitioned into two contiguous  folds of approximately equal length: fold~1 covers samples
$6$--$515$ and fold~2 covers samples $516$--$1024$. For each fold, the initial-condition vector is constructed
by stacking the five most recent input and output samples immediately
preceding it.  
At each round, one fold serves as the validation set while the other is
used for model training.
The cross-validation RMSE, computed as the average open-loop simulation
RMSE over the two folds, constitutes the  metric $V(\theta)$ (eq.~\eqref{eq:cv_metric}) used by \name{} to rank candidate configurations. 

\subsubsection{Agent search results}

\begin{figure}
\begin{center}
\includegraphics[width=8.4cm]{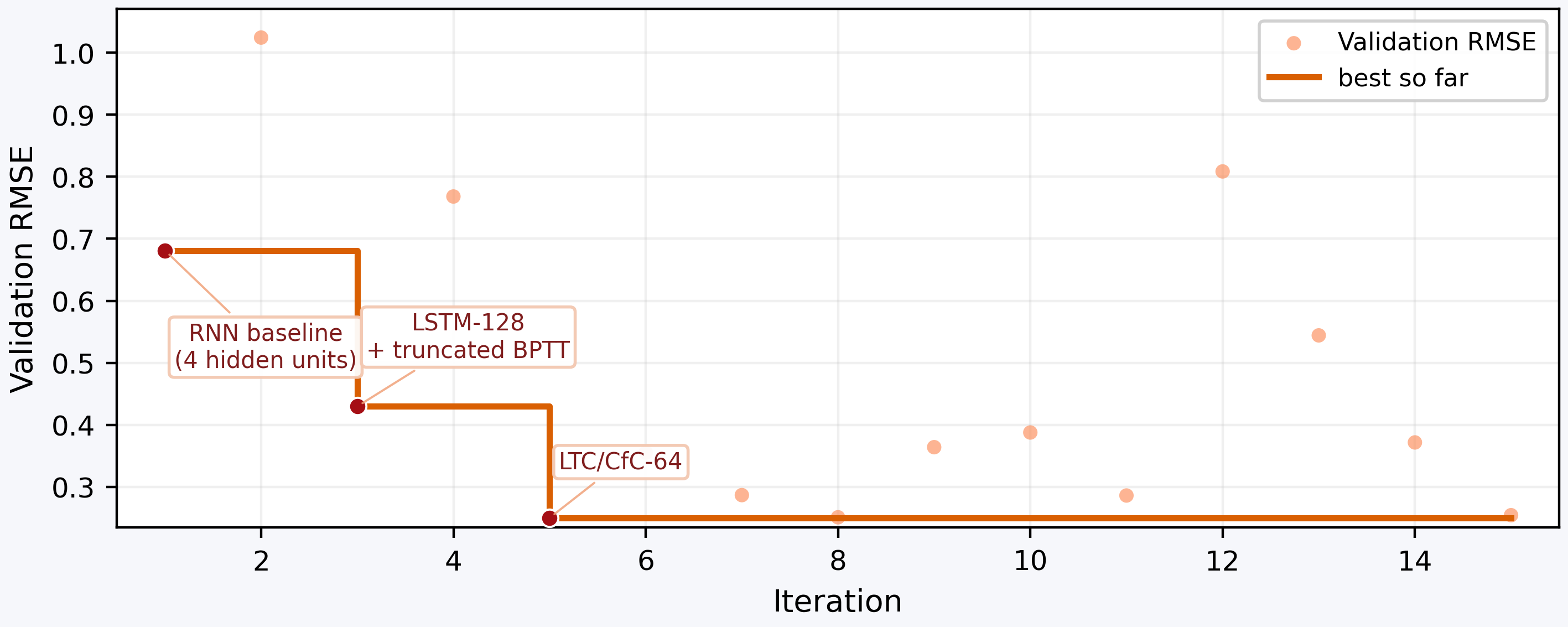}    
\caption{Cascaded tanks: Normalized validation RMSE \emph{vs} iterations of \name{}. RMSE obtained at each iteration (point) and  best value achieved up to that iteration (line). Annotated tags mark the main modifications introduced where a new best is reached.} 
\label{fig:cascaded_final_analysis}
\end{center}
\end{figure}

The agentic framework explored 15 candidate configurations over the course of the search,
producing three successive improvements over the initial baseline. Evolution of the normalized validation RMSE over the iterations is visualized in Fig.~\ref{fig:cascaded_final_analysis}. 
The starting model is a vanilla RNN with only four hidden
units. It achieved a validation RMSE of $0.681$. At iteration~3, the agent hypothesised model capacity as the primary bottleneck and replaced the architecture with a two-layer LSTM with 128 hidden units. Furthermore, the training algorithm was also modified by the agent into a truncated backpropagation through time algorithm (chunk size 50 steps). This reduced the validation RMSE to 0.430. 
At iteration~5, the agent introduced an inductive bias, replacing the LSTM with a Liquid Time-Constant / Closed-form Continuous-time (LTC/CfC) network~\cite{Hasani2022}
with $n_h = 64$ hidden states. This architecture constitutes a strong inductive  bias for dissipative systems such as the cascaded tank. In fact, unlike a generic LSTM,
the model is structurally predisposed to solutions that decay over time, which matches the physical behaviour of~\eqref{eq:tank_ode} even without any explicit knowledge of the parameters $k_1,\ldots,k_4$. The result was a substantial reduction in validation RMSE, reaching $0.250$.

The remaining ten iterations explored a range of alternative configurations,
including grey-box ODE models, dilated causal CNNs, cascaded two-timescale
LTC networks, diagonal state-space models, and GRU-based architectures.
None of these produced a further reduction in validation RMSE; the LTC/CfC
model with $n_h = 64$ hidden states was therefore retained as the final model.

\begin{figure}[t]
    \centering
    \includegraphics[width=0.49\columnwidth]{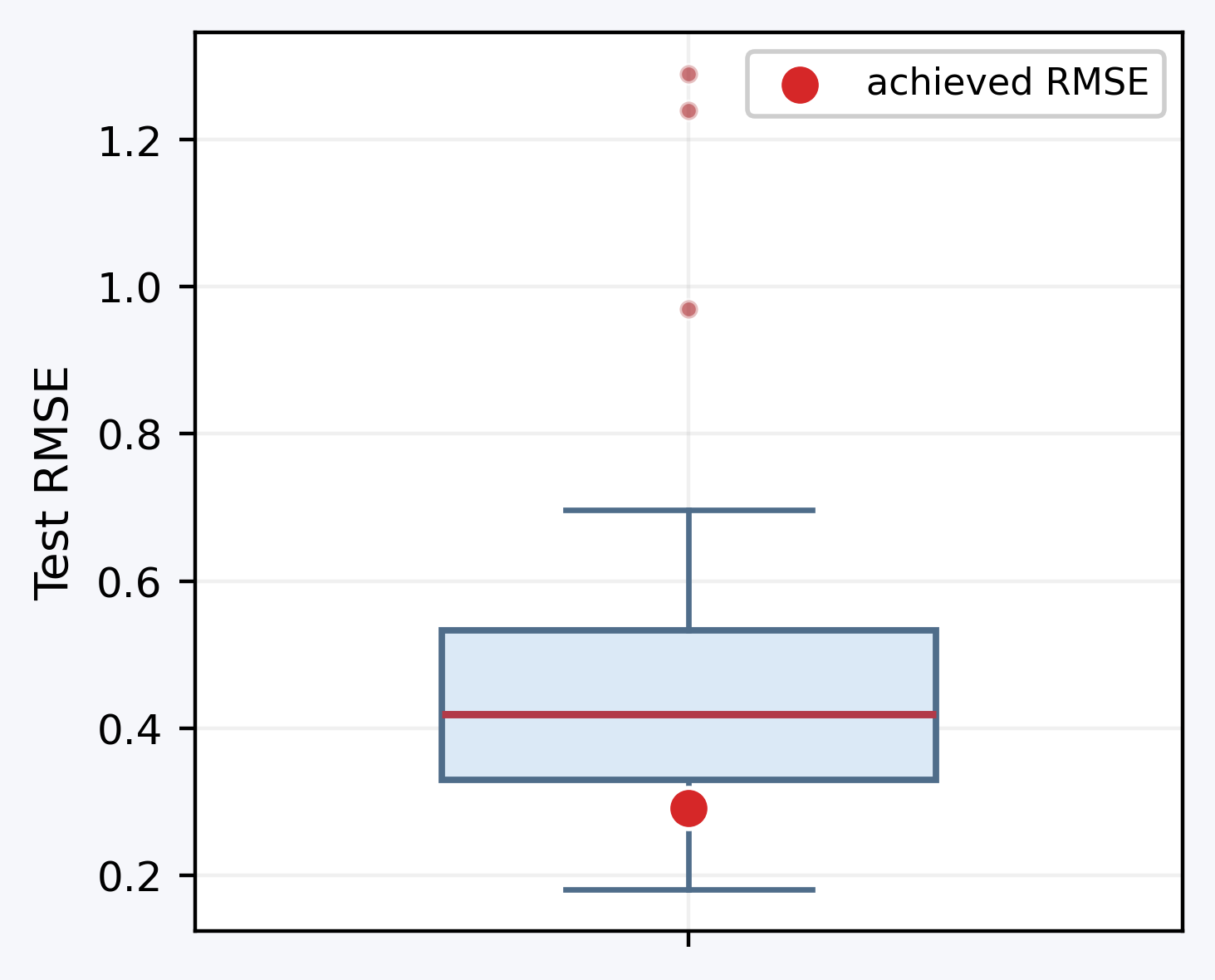}%
    \hfill%
    \includegraphics[width=0.49\columnwidth]{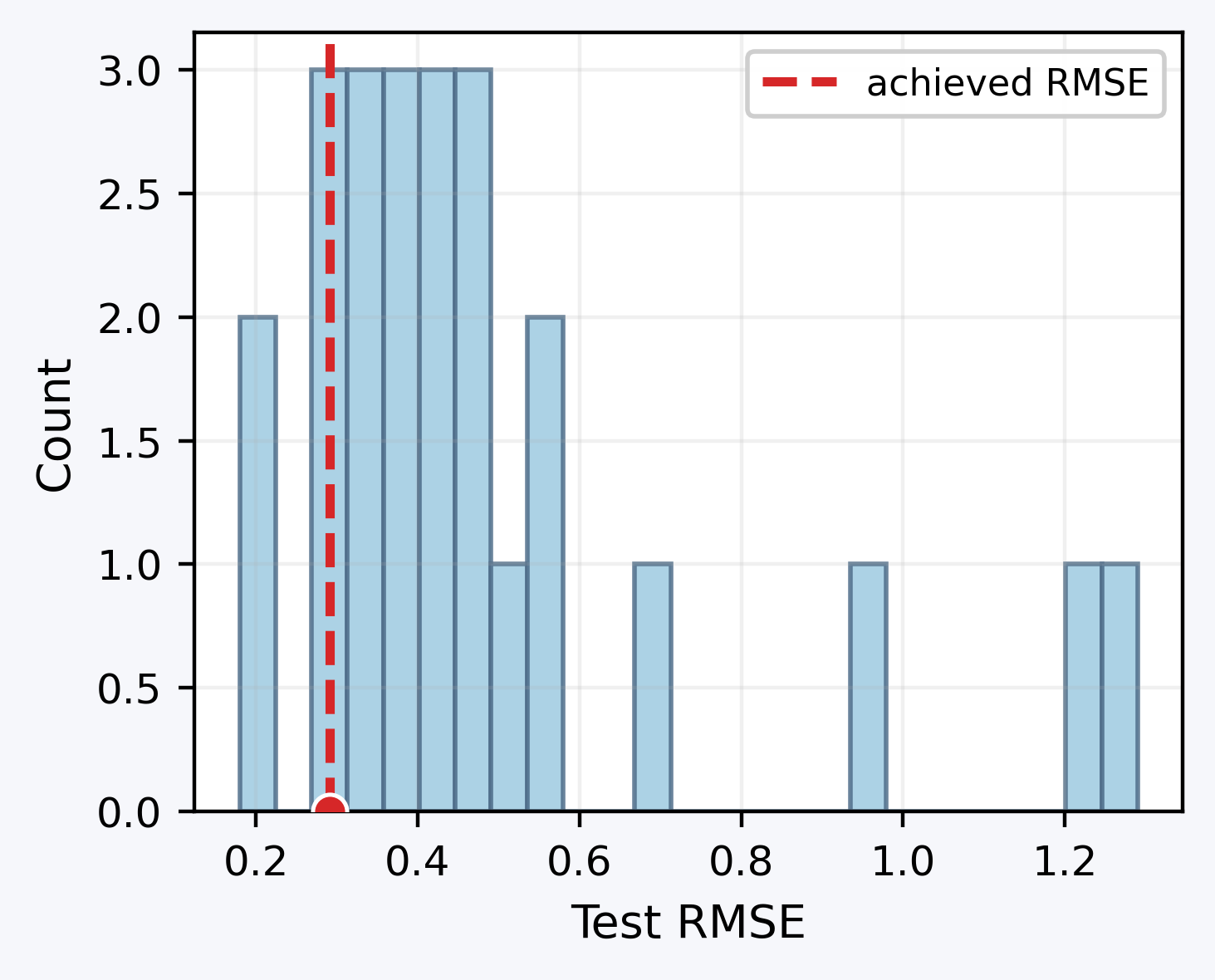}
    \caption{Cascaded tanks: distribution of test RMSE values reported
         in the benchmark leaderboard, shown as box-plot (left)  and histogram (right). Red marker and dashed line indicate
         the test RMSE achieved by \name{}.}
    \label{fig:boxplot_histogram}
\end{figure}

The final model used for testing  is an ensemble of
two instances of the trained  LTC/CfC
 models  (one per cross-validation fold), with predictions averaged
at inference time. The   RMSE achieved  in the  test trajectory   amounts to $0.298$ (denormalized units).
A comparison with results reported in the benchmark leaderboard\footnote{\url{https://www.nonlinearbenchmark.org/benchmarks/cascaded-tanks.} Accessed: 2026-05-06} 
is shown in Fig.~\ref{fig:boxplot_histogram}, which displays
the box plot and histogram of the test RMSE values available in the leaderboard.
As can be observed, the achieved result is in line with the best results
reported in the literature.

\subsection{Nanodrone}

\subsubsection{Benchmark description}
\noindent
The second benchmark concerns the identification of the flight dynamics of a
Crazyflie~2.1 brushless nano-quadrotor~\cite{Busetto26}.
The platform is a four-propeller nanodrone actuated by four brushless motors.
The system input $u = [u_1, u_2, u_3, u_4]^\top \in \mathbb{R}^4$ collects
the propeller rotational speeds.
The full state vector
\[
  x = \bigl[p^\top,\;v^\top,\;\Phi^\top,\;\omega^\top\bigr]^\top \in \mathbb{R}^{12}
\]
comprises 3D position coordinates $p=[p_x,p_y,p_z]^\top$, linear velocity
$v=[v_x,v_y,v_z]^\top$, attitude expressed as Euler angles
$\Phi=[\phi,\theta,\psi]^\top$, and body-frame angular
velocity $\omega=[\omega_x,\omega_y,\omega_z]^\top$.
All 12 state components are  measured.

The nominal dynamics of the quadrotor follow the nonlinear rigid-body ODEs:
\begin{equation}
  \left\{
  \begin{aligned}
    \dot{p}    &= v,\\[2pt]
    m\,\dot{v} &= R(\Phi)\begin{bmatrix}0 \\ 0 \\ k_T\displaystyle\sum_{i=1}^4 u_i^2\end{bmatrix}
                  - m g\,e_z,\\[6pt]
    \dot{\Phi} &= T(\Phi)\,\omega,\\[2pt]
    J\,\dot{\omega} &= \tau(u) - \omega\times J\omega,
  \end{aligned}
  \right.
  \label{eq:drone_ode}
\end{equation}
where $R(\Phi)$ is the body-to-inertial rotation matrix,
$T(\Phi)$ the Euler-rate kinematic matrix, $m$ the drone mass, $J$ the
moment-of-inertia tensor, $g$ gravitational acceleration,
$e_z=[0,0,1]^\top$, and $k_T$ the propeller thrust coefficient.
The generalised torque $\tau(u)$ is a function of motor speed differences
weighted by the arm length $l$ and the drag-to-thrust ratio. 
The nominal model~\eqref{eq:drone_ode} neglects blade-flapping,
rotor inertia, aerodynamic interference between propellers, and structural
flexibility, all of which become significant during aggressive manoeuvres.

The dataset consists of real-world flight recordings sampled at
$T_s = 10$\,ms across four distinct trajectories:
\emph{chirp} (sinusoidally swept-frequency excitation), \emph{random}
(pseudo-random motor commands), \emph{square} (step-like commands producing
sharp direction reversals), and \emph{melon} (elliptical trajectories replicating the surface lines of a
netted melon). 
The chirp, random, and square trajectories, each comprising approximately
$18\,500$ samples, are available for training and cross-validation; while the melon
trajectory (approximately
$19\,500$ samples) is reserved  for  final testing.

\subsubsection{Evaluation criterion}
\noindent
A leave-one-trajectory-out cross-validation scheme is adopted to guide
the agentic search.
The three training trajectories define three folds: for each fold, one
trajectory serves as the validation set while the remaining two are used
for training.  For each fold, the held-out validation trajectory is partitioned into non-overlapping
windows of $H=50$ steps (${\approx} 0.5$\,s).
For each window, the model is initialised from the true state at the previous 
sample and rolled out   for 50 steps with no access to
intermediate ground-truth states.
Validation performance  is quantified by the mean absolute error (MAE) averaged over
all prediction steps, all output channels, and all windows:
\begin{equation}
  \mathrm{MAE} =
  \frac{1}{N\,H\,n_y}
  \sum_{j=1}^{N}\sum_{k=1}^{H}\sum_{i=1}^{n_y}
  \bigl|\tilde{x}_{j,k,i} - \hat{x}_{j,k,i}\bigr|,
  \label{eq:drone_mae}
\end{equation}
where $n_y=12$, $H=50$, $N$ is the number of windows, $\tilde{x}_{j,k,i}$
is the $i$-th normalised state component at step $k$ of window $j$, and
$\hat{x}_{j,k,i}$ the corresponding model prediction.
State variables are normalised channel-wise
so that outputs with different physical scales contribute equally to the
metric.
The aggregate cross-validation MAE, i.e.\ the arithmetic mean of the three
per-fold values, is the sole criterion $V(\theta)$ (eq.~\eqref{eq:cv_metric}) used by \name{} to rank
candidate configurations.

\subsubsection{Agent search results}

\begin{figure}
\begin{center}
\includegraphics[width=8.4cm]{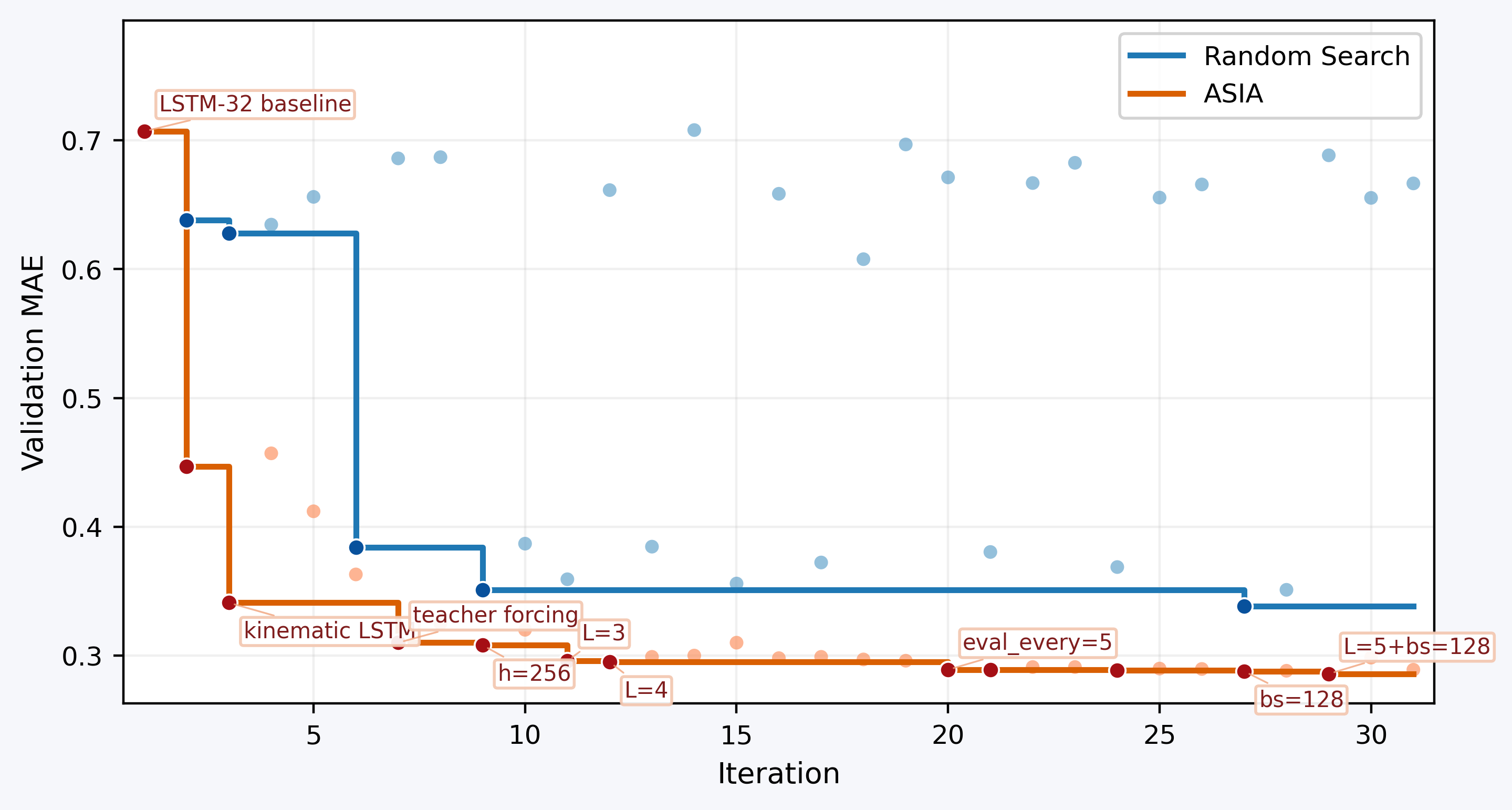}
\caption{Nanodrone: comparison between random search (blue) and agentic AI
(orange) in terms of validation MAE \emph{vs} iterations. MAE at each
iteration (point) and best value achieved up to that point (solid lines).}
\label{fig:drone_final_analysis}
\end{center}
\end{figure}

The starting model is a single-layer LSTM with 32 hidden units trained with
mean-squared-error loss, achieving a validation MAE of $0.707$. \name{} evaluated 31 candidate configurations in total
(baseline plus 30 iterations), producing eight successive improvements.

As a baseline, a random search over the same budget of 30
iterations was conducted. Each configuration was drawn by randomly sampling
the model class (pure autoregressive LSTM or physics-residual LSTM),
the hidden state dimension, the number of layers, the learning rate, the dropout rate, and the weight decay. 
The evolution of the cross-validation MAE for both search strategies is
visualized in Fig.~\ref{fig:drone_final_analysis}, which shows that
\name{} outperforms random search both in terms of the best validation
MAE achieved and in terms of search efficiency. In particular, after only
three iterations, \name{} already reaches the best performance obtained
by random search over the entire search budget.

 For the \name{} framework, at iteration~3 the agent introduced a
physics-informed architectural restructuring. In particular, the baseline LSTM is replaced by   a custom architecture (referred to as \emph{KinematicsLSTM}\footnote{Name given by the AI Agent.} in Fig.~\ref{fig:drone_final_analysis}), in which
 the recurrent
network is tasked exclusively with predicting velocity and angular-velocity
increments $(\Delta v, \Delta\omega)$, while position and attitude are
recovered via explicit kinematic integration,
\begin{align}
      p_t = p_{t-1} + \gamma_p \odot v_{t-1}, \qquad
  \Phi_t = \Phi_{t-1} + \gamma_\Phi \odot \omega_{t-1},
\end{align}
where $\odot$ denotes point-wise multiplication, and  $\gamma_p,\gamma_\Phi\in\mathbb{R}^3$ 
are  learnable   gains.
Furthermore, the motor-speed inputs were augmented with their element-wise
squares $[u_1^2,u_2^2,u_3^2,u_4^2]$, providing the network with a direct
proxy for aerodynamic thrust. 
At iteration~7 the agent introduced scheduled sampling to stabilise training. Since the recurrent model is rolled out using its own predictions at every training step, errors accumulate, especially early in training, and may lead to unstable gradient descent. 
Injecting the ground-truth state during training with a probability that decays linearly
from $0.3$ to $0$ over the training epochs mitigates this instability, reducing the validation MAE to $0.310$.

Subsequent iterations increased model capacity by widening the hidden state
and stacking additional LSTM layers. Other   improvements were obtained by refining the training procedure, such as a finer model evaluation cadence to improve the  early-stopping criterion, and a larger batch size. This yields   a final best cross-validation MAE of $0.286$. 
Throughout the search, several alternative directions were also explored, 
including Koopman operator models, causal Transformers, GRU-based architectures, input noise augmentation, dropout, and weight decay tuning.

\subsubsection{Test results} 
Like in the cascaded two-tank benchmark, the final model is an ensemble of
three instances (one per cross-validation fold). To enable direct comparison with~\cite{Busetto26}, the test evaluation
adopts the same metric definitions in~\cite{Busetto26}. In particular, for each output group $g \in \{p,\,v,\,R,\,\omega\}$ (positions, linear
velocities, attitudes, and angular velocities), the \emph{Mean Euclidean
Error}  (MEE) at horizon~$h$ is computed, i.e., 
\begin{equation}
  \mathrm{MEE}_{g,h}
  = \frac{1}{N}\sum_{j=1}^{N}
    \bigl\|x^{(g)}_{j,h} - \hat{x}^{(g)}_{j,h}\bigr\|_2,
  \label{eq:mee}
\end{equation}
where $N$ is the number of test windows and
$x^{(g)}_{j,h},\hat{x}^{(g)}_{j,h}\in\mathbb{R}^3$ are the true and
predicted state components of group~$g$ at step~$h$ of window~$j$.
For positions, velocities, and angular velocities, $\|\cdot\|_2$ denotes
the standard Euclidean norm. For attitudes, in line with~\cite{Busetto26},
the geodesic distance   is used instead.
The cumulative error $\Sigma = \sum_{h=1}^{50}\mathrm{MEE}_{g,h}$
summarises accuracy over the full prediction horizon.

Numerical results are reported in Table~\ref{tab:drone_horizons}, which
also includes a na\"ive constant predictor as a lower-bound reference.
The model found by \name{}  outperforms both the best method
of~\cite{Busetto26} (physics+residual network) and the random search
baseline across all output groups.  

\begin{table*}
\caption{Nanodrone: Mean Euclidean Error (MEE) per output group at horizons
$h=10$, $50$ and cumulative ($\Sigma_{h=1}^{50}$), evaluated on the
held-out Melon test trajectory. \textbf{Bold}: best per column.
Models compared: na\"ive predictor; physics+residual network
from~\cite{Busetto26}; best model found by random search; best model
found by \name{}.}
\label{tab:drone_horizons}
\centering
\begin{tabular}{l
  rrr @{\hspace{6pt}}
  rrr @{\hspace{6pt}}
  rrr @{\hspace{6pt}}
  rrr}
\toprule
& \multicolumn{3}{c}{$\mathrm{MEE}_{p,h}$ [m]}
& \multicolumn{3}{c}{$\mathrm{MEE}_{v,h}$ [m/s]}
& \multicolumn{3}{c}{$\mathrm{MEE}_{R,h}$ [rad]}
& \multicolumn{3}{c}{$\mathrm{MEE}_{\omega,h}$ [rad/s]} \\
\cmidrule(lr){2-4}\cmidrule(lr){5-7}
\cmidrule(lr){8-10}\cmidrule(lr){11-13}
Model
  & $h{=}10$ & $h{=}50$ & $\Sigma$
  & $h{=}10$ & $h{=}50$ & $\Sigma$
  & $h{=}10$ & $h{=}50$ & $\Sigma$
  & $h{=}10$ & $h{=}50$ & $\Sigma$ \\
\midrule
Na\"ive
  & 0.143 & 0.680 & 17.8
  & 0.318 & 1.475 & 38.9
  & 0.069 & 0.304 &  8.2
  & 0.360 & 0.884 & 29.1 \\
Phys+Res~\cite{Busetto26}
  & 0.017          & 0.112          &  2.4
  & \textbf{0.061} & 0.556          & 10.4
  & 0.038          & 0.231          &  6.2
  & 0.488          & 0.598          & 29.0 \\
Random search
  & 0.015          & 0.124          &  2.6
  & 0.078          & 0.417          & 10.1
  & 0.026          & 0.155          &  4.3
  & 0.354          & 0.581          & 22.4 \\
\name{}
  & \textbf{0.013} & \textbf{0.096} & \textbf{2.1}
  & 0.077          & \textbf{0.342} & \textbf{8.4}
  & \textbf{0.027} & \textbf{0.149} & \textbf{3.9}
  & \textbf{0.314} & \textbf{0.450} & \textbf{19.6} \\
\bottomrule
\end{tabular}
\end{table*}

\section{Discussion and Limitations} \label{sec:discussion}

This section discusses the limitations and practical considerations arising from our experience in developing and using \name{}.

\subsection{Test  Leakage}

In our experiments, the agent was not allowed to use the test set for model selection or hyperparameter tuning.  
However, test performance was logged to monitor progress.
While \name{} was explicitly not allowed  to use this information, the AI agent  may still be influenced  indirectly through its  reasoning process. 
This phenomenon reflects a broader issue  present in  machine learning research. 
Although the test set is formally reserved for final evaluation, the iterative nature of model development (where practitioners repeatedly refine architectures based on observed performance) can lead to implicit forms of test-driven reasoning.
In practice, even careful researchers may be influenced by test results when deciding whether a model is satisfactory or when selecting the best identification strategy.

A related and more general source of implicit test information leakage in \name{} arises from prior knowledge.
When the agent is provided with references to the literature, the selection of model classes and architectures may be indirectly guided by published results, which themselves are typically reported on test datasets.
As a consequence, part of the information contained in the test distribution may be implicitly encoded in the prior knowledge available to the agent.

\subsection{Training Pipeline as a Black Box}

A key feature of \name{} is that it  modifies  code and learning strategies.  
It is thus important to distinguish two levels of interpretability.
At the local level, the generated code is, in principle, interpretable, as each function and modification can be inspected and verified.
However, at the global level, the sequence of decisions that led to the final configuration is not easily analysable, as it is   by construction iteratively automatically generated. 
This aspect is particularly relevant in  research.
Traditionally, the scientific contribution lies not only in the final model, but in the methodology that justifies its design.
When the training procedure itself emerges from an automated and partially opaque search process, this methodological contribution becomes less explicit, raising questions about interpretability, reproducibility, and scientific insight.

\subsection{Model versus Methodology}

In practical applications, the end user is typically interested in the final model, regardless of how it was obtained.
From this perspective, \name{} successfully produces competitive models, often combining physical structure with black-box components, as observed  in the two  benchmarks.  
However, as discussed in the previous paragraph,  the primary output for a system identification researcher is  not only the model, but also the identification methodology.
The limited interpretability and reproducibility of automatically generated training strategies may therefore reduce their scientific value.

This highlights a conflict  between practical  effectiveness of the final result and methodological transparency.
While \name{} may accelerate the discovery of high-performing models, it may simultaneously obscure the underlying principles that would traditionally constitute the main contribution in system identification research.

\subsection{Exploration of the Hypothesis Space}

\name{} introduces  significant benefits  in terms of exploration.
By autonomously generating  alternative modelling and training strategies, the agent effectively explores a rich \emph{hypothesis space}, which may be extremely large and, in principle, unbounded.
 In contrast, classical approaches such as grid search, evolutionary algorithms, or Bayesian optimisation  operate over predefined, finite-dimensional search spaces (namely, $\mathbb{R}^n$, where $n$ is the number of tunable parameters). 
As a consequence, they are inherently limited to variations of a fixed model class and parameterisation. 

In our experiments, the agent was able to propose non-trivial modelling choices and training strategies, including hybrid physical/black-box structures, Koopman-inspired representations, reservoir computing architectures, and alternative training schemes such as multi-horizon losses or teacher forcing.
These choices were not explicitly specified a priori, but emerged from the agent's iterative reasoning process. This capability suggests that the value of \name{} is not limited to automation, but also lies in its ability to expand the space of candidate solutions and stimulate new research directions.

\subsection{Bias and Reproducibility}

The search process induced by LLMs is inherently biased by
its training data and internal priors. As a result, \name{} may favour
well-established modelling choices, architectures, or training strategies
over less conventional alternatives. 

Reproducibility raises a second issue. Different executions of the
same agentic search may lead to different model architectures,
hyperparameters, and training strategies.  This is due to the stochastic
nature of LLM outputs. Therefore,
the search trajectory followed by an AI agent can not be 
reproduced exactly. This limitation can be  mitigated by using LLMs under full user
control, such as open-weight models operated locally. In such
settings, model weights, decoding parameters, prompts,  and random seeds can be fixed and documented, making the agentic search   reproducible.

It is important, however, to distinguish the reproducibility of the search
process from that of the final identified model. Once a model class,
architecture, hyperparameters, training code,  and random seeds are
fixed, the LLM is no longer involved. At that stage, the training and
evaluation pipeline reduces to a conventional training algorithm, which  can be reproduced using standard practices. Thus, the main reproducibility concern
does not concern the final model per se, but the automated process that led to
its selection.

\section{Conclusion} \label{sec:conclusion}

This paper presented \emph{\name{}}, an agentic AI framework for system
identification   acting as an autonomous coding
agent.  The empirical results demonstrate that the proposed framework can successfully automate substantial parts of the system identification pipeline and discover competitive solutions, including non-trivial  architectures and training strategies. At the same time, our study also highlights important limitations, including implicit biases and the limited transparency of the resulting training proces.

Overall, these observations suggest that \name{} should not be viewed as a
replacement for human-driven system identification, but rather as a
complementary tool. Its primary value lies in accelerating experimentation,
automating parts of the design process, and expanding the exploration of the
hypothesis space. Human expertise remains essential not only for interpreting
results and validating the final models, but also for designing 
prompts, injecting relevant prior knowledge, and guiding the agent toward
promising modelling directions.

More broadly, this work aims to provide an initial study and step toward the 
usage of agentic AI in system identification  and related engineering
areas. Similar ideas could be explored for
automatic controller tuning, including PID and MPC hyperparameter optimisation,
the design of hybrid model-based and learning-based controllers, and experiment
design, just to cite a few. 
Future research may also investigate integration with formal verification tools, in order to provide theoretical  guarantees of the automatically generated  pipelines.


\section*{DECLARATION OF AI-ASSISTED TECHNOLOGIES IN THE WRITING PROCESS}

This work explicitly investigates the use of generative AI technologies for  research. The proposed
framework is itself based on autonomous AI agents used to generate, modify, and execute training code
during the experiments  reported in the paper. In addition, during the preparation of the manuscript, the authors used ChatGPT to support  text refinement.
However, the overall research questions, experimental design, and methodological structure  were conceived  by the authors. All content generated with the support of generative AI was reviewed, validated, and edited by the authors.

\bibliographystyle{IEEEtran}
\bibliography{references}

\end{document}